\documentclass[letterpaper, 12pt]{article}

\usepackage{multicol}
\usepackage{booktabs}
\usepackage{longtable}
\usepackage{amssymb,amsmath}
\usepackage{amstext}
\usepackage[bookmarks=true,bookmarksnumbered=true]{hyperref}
\usepackage[square,sort]{natbib}
\usepackage{bm}
\usepackage{tikz}
\usepackage{epstopdf}
\usepackage{float}
\usepackage{subfig}
\usepackage[titlenumbered,ruled]{algorithm2e}
\usepackage[font=small,skip=0pt]{caption}

\usetikzlibrary{shapes,arrows}

\usepackage{qresize}
\resize

\if@symbold\else\fi

\newcommand{\x}{{\bf x}}
\newcommand{\X}{{\bf X}}
\newcommand{\y}{{\bf y}}
\newcommand{\R}{{\bf R}}
\newcommand{\W}{{\bf W}}
\newcommand{\w}{{\bf w}}
\newcommand{\p}{{\bf p}}
\newcommand{\e}{{\bf e}}

\SetCommentSty{mycommfont}

\usepackage{pdfsync}

\title{Gradient-enhanced kriging for high-dimensional problems}
\author{Mohamed A. Bouhlel \\ \href{mailto:mbouhlel@umich.edu}{mbouhlel@umich.edu}
\and Joaquim R. R. A. Martins \\ \href{mailto:jrram@umich.edu}{jrram@umich.edu}
}

\begin{document}
\maketitle

\begin{abstract}

Surrogate models provide a low computational cost alternative to evaluating expensive functions.
The construction of accurate surrogate models with large numbers of independent variables is currently prohibitive because it requires a large number of function evaluations.
Gradient-enhanced kriging has the potential to reduce the number of function evaluations for the desired accuracy when efficient gradient computation, such as an adjoint method, is available.
However, current gradient-enhanced kriging methods do not scale well with the number of sampling points due to the rapid growth in the size of the correlation matrix where new information are added for each sampling point in each direction of the design space.
They do not scale well with the number of independent variables either due to the increase in the number of hyperparameters that needs to be estimated.
To address this issue, we develop a new gradient-enhanced surrogate model approach that drastically reduced the number of hyperparameters through the use of the partial-least squares method that maintains accuracy.
In addition, this method is able to control the size of the correlation matrix by adding only relevant points defined through the information provided by the partial-least squares method.
To validate our method, we compare the global accuracy of the proposed method with conventional kriging surrogate models on two analytic functions with up to 100 dimensions, as well as engineering problems of varied complexity with up to 15 dimensions.
We show that the proposed method requires fewer sampling points than conventional methods to obtain a desired accuracy, or provides more accuracy for a fixed budget of sampling points.
In some cases, we get over 3 times more accurate models than a bench of surrogate models from the literature, and also over 3200 times faster than standard gradient-enhanced kriging models.
\end{abstract}

\newpage
\section*{Symbols and notation}
Matrices and vectors are in bold type.
\begin{longtable}{ll}
\endhead  
\begin{tabular}{ll}
\bf{Symbol} & \bf{Meaning}\\[8pt]
$d$ & Number of dimensions\\
$B$ & Hypercube expressed by the product between intervals of each direction space\\
$n$& Number of sampling points\\
$h$& Number of principal components\\
$\x$, $\x'$& $1\times d$ vector\\
$x_j$& $j^{\text{th}}$ element of $\x$ for $j=1,\dots,d$\\
$\X$& $n\times d$ matrix containing sampling points\\
$\y$& $n\times 1$ vector containing simulation of $\X$\\
$\x^{(i)}$& $i^{\text{th}}$ sampling point for $i=1,\dots,n$ ($1\times d$ vector)\\
$y^{(i)}$& $i^{\text{th}}$ evaluated output point for $i=1,\dots,n$\\
$\X^{(0)}$&$\X$\\
$\X^{(l-1)}$& Matrix containing residual of the $(l-1)^{\text{th}}$ inner regression\\
$k(\cdot,\cdot)$& Covariance function\\
${\bf r}_{\x\x'}$& Spatial correlation between $\x$ and $\x'$\\
$\R$& Covariance matrix\\
$s^2(\x)$& Prediction of the kriging variance\\
$\sigma^2$& Process variance\\
$\theta_i$& $j^{th}$ parameter of the covariance function for $i=1,\dots,d$\\
$Y(\x)$& Gaussian process\\
${\bf 1}$& $n$-vector of ones\\
${\bf t}_l$& $l^{th}$ principal component for $l=1,\dots,h$\\
${\bf w}$& Weight vector for partial least squares\\
$\Delta x_j$& First order Taylor approximation step in the $j^{th}$ direction\\
\end{tabular}
\endfoot
\end{longtable}

\section{Introduction}
\label{sec:intro}

Surrogate models, also known as metamodels or response surfaces, consist in approximate functions (or outputs) over a space defined by independent variables (or inputs) based on a limited number of function evaluations (or samples).
The main motivation for surrogate modeling is to replace expensive function evaluations with the surrogate model itself, which is much less expensive to evaluate.
Surrogate model approaches often used in engineering applications include polynomial regression, support vector machine, radial basis function models, and kriging~\citep{Forrester2008}.
Surrogate models are classified based on whether they are non-interpolating, such as polynomial regression, or interpolating, such as kriging.
Surrogate models can be particularly helpful in conjunction with numerical optimization, which requires multiple function evaluations over a design variable space~\citep{Haftka2016,Jones2001,Simpson2001a}.
However, non-interpolating surrogate models are not sufficient to handle optimization problems because adding additional points does not necessarily lead to a more accurate surface~\citep{Jones2001}.
On the other hand, interpolating surrogate models become accurate in a specific area where new points are added.
One of the most popular interpolating models is the kriging model~\citep{Krige1951,Matheron1963,Cressie1988,Sacks1989a,Simpson2001b}, also known as Gaussian process regressions \citep[Ch. 3, Sec. 19]{Barber2012,Rasmussen2006}.
\cite{Kleijnen2009} gives a general review of kriging, and presents the basic assumptions and formulas.
Compared to other common surrogate models, one of the major advantages of kriging is the built-in analytical estimate of the model error, which makes kriging a probabilistic model for which we can use statistical techniques~\citep{Jones1998}.
Several efforts have shown that kriging can significantly reduce the cost of numerical analysis and optimization.
\cite{Jeong2005}, for example, used a kriging to model a two-dimensional airfoil design including flap position in a multi-element airfoil, where the lift-to-drag ratio was maximized using a genetic algorithm.
Since genetic algorithms require a large number of function evaluations, the kriging surrogate greatly reduced the overall computational cost.
\cite{Toal2008} used two kriging-based optimizations with an intermediate step that uses a proper orthogonal decomposition method to minimize the drag-to-lift ratio of a 2-$d$ transonic airfoil design.
This approach outperformed a traditional kriging-based optimization, producing better designs for a considerable reduction of the optimization cost, and this was done by restricting the budget of the first optimization and by keeping only the relevant points for the second optimization.
\cite{Nathalie2016}, for example, used a mixture of experts involving several types of kriging to minimize the weight of an internal aircraft wing structure.
The structural optimization considered 12 thickness variables (spars, skins and ribs) and 2 stress constraints (spars and skins).
Their results showed that optimization based on the kriging models required fewer evaluations than a direct optimization method.
Many other applications using the kriging model could be found in the literature~\citep{Sakata2003,Kleijnen2010,Kleijnen2012,Liem2015,Choi2004,Liem2012a,Liem2015a}.

Kriging models can be extended to utilize gradient information when available, which improves the accuracy of the model.
Such methods are known in the literature as gradient-enhanced kriging (GEK)~\citep{Liem2015}, cokriging~\citep{Laurenceau2008,chung2002}, or first-order kriging~\citep{Lewis1998}.
GEK has been shown to be effective in various studies~\citep{Laurenceau2008,chung2002,Lewis1998,Liu2003}, and are especially advantageous when the gradient is computed with an adjoint method, where the cost of computing the gradient is independent of the number of independent variables~\citep{Martins2013a}.
\cite{Laurenceau2008} compared kriging and direct-indirect GEK (using a discrete adjoint method for computing the gradients) and showed a considerable gain in global accuracy using the indirect GEK on an aerodynamic shape optimization problem.
Despite this performance, the number of input variables was still low (2 to 6) because of the exorbitant computational cost required to build GEK for larger inputs.
\cite{Liem2015} used a mixture of experts method using GEK to approximate the drag coefficients on a surrogate-based aircraft mission analysis.
This method is compared to conventional surrogate models showing the superiority of GEK models, especially in terms of accuracy.
Similarly to \cite{Laurenceau2008}, the number of input variables was low (2 and 4).

GEK is subject to performance degradation when the number of input variables, the number of sampling points, or both, are high.
This performance degradation is mainly due to the size of the GEK correlation matrix, which increases proportionally with both the number of inputs and the number of sampling points.
In addition, when sampling points are close to each other, this leads to quasi-linearly dependent columns in the correlation matrix that makes it ill-conditioned, and the corresponding linear problem becomes challenging to solve.
There are other difficulties in high-dimensional problems because building a kriging surrogate model involves solving a multimodal optimization problem whose number of variables is proportional to the problem dimension.
This optimization problem involves maximizing a function---the \emph{likelihood function}---with respect to variables called \emph{hyperparameters}.

To address the difficulty in finding the hyperparameters through optimization, \cite{Laurenceau2008} developed a method that guesses an initial solution of the GEK hyperparameters, and then uses a gradient-based optimization method to maximize the likelihood function.
This method accelerates the construction of the GEK model; however, the initial guess depends on a fixed parameter that defines the strength of the correlation between the two most directional-distant sample points.
This fixed parameter depends on the physical function to be studied and thus requires trial and error.
Therefore, it is not easy to generalize this approach.
\cite{Lewis1998} also tried to accelerate the estimation of the GEK hyperparameters by reducing their number to one for all directions.
The GEK model has shown better results than conventional kriging (using one hyperparameter for all directions) on a borehole flow-rate problem using 8 input variables.
However, they assumed that the problem is isotropic, which is not the case for the engineering problems we want to tackle.

\citet{Bouhlel2016a} proposed an approach that consists in combining the kriging model with the partial-least squares (PLS) method, called KPLS, to accelerate the kriging construction.
This method reduces the number of the kriging hyperparameters by introducing new kernels based on the information extracted from the PLS technique.
The number of hyperparameters is then reduced to the number of principal components retained.
Experience shows that 2 or 3 principal components are usually sufficient to get good accuracy~\citep{Bouhlel2016a}.
There is currently no rule of thumb for the maximum number of principal components to be retained because it depends of both the problem and location of the sampling points used to fit the model.
The KPLS model has shown to be efficient for several high-dimensional problems.
\cite{Bouhlel2016a} compared the accuracy between KPLS and conventional kriging models on analytical and engineering problems problems with a number of inputs up to 100.
Despite the reduced number of hyperparameters used into the KPLS model, they obtained similar results in terms of accuracy between both models.
The main benefit of KPLS was a reduction in the computational time needed to construct the model: KPSL was up to 450 faster than conventional kriging.

Another variant of KPLS, called KPLSK, was also developed by \citet{Bouhlel2016b} that extends the KPLS method by adding a new step into the construction of the model.
Once the KPLS method is built, the hyperparameters' solution is considered as a first guess for a gradient-based optimization applied on a conventional kriging model.
The idea of the KPLSK method is similar to that developed by \citet{Ollar2016}, where a gradient-free optimization algorithm is used with an isotropic kriging model followed by a gradient-based optimization starting from the solution provided by the first optimization.
The results of KPLSK have shown a significant improvement on analytical and engineering problems with up to 60 dimensions in terms of accuracy when compared to the results of KPLS and conventional kriging.
In addition, the KPLSK model is more efficient than kriging (up to 131 faster using 300 points for a 60$D$ analytical function), and, however, is slightly less efficient than KPLS (22 s vs 0.86 s, respectively, for KPLSK and KPLS with the same test case).
An optimization applications using KPLS and KPLSK could be found in the literature \citep{Nathalie2016,Bouhlel2016c}.

To further improve the efficiency of KPLS and extend GEK to high-dimensional problems, we propose to integrate the gradient during the construction of KPLS and a different way to use the PLS method.
This approach is based on the first order Taylor approximation (FOTA) at each sampling point.
Using this approximation, we generate a set of points around each sampling point and apply the PLS method for each of these sets.
We then combine the information from each set of points to build a kriging model.
We call this new model GE-KPLS since such construction uses both the gradient information and the PLS method.
The GE-KPLS method utilizes gradient information and controls the size of the correlation matrix by adding some of the approximating points in the correlation matrix with respect to relevant directions given by the PLS method at each sampling point.
The number of hyperparameters to be estimated remains equal to the number of principal components.

The remainder of the paper is organized as follows.
First, we review the key equations for the kriging and KPLS models in Sections~\ref{sec:krig} and~\ref{sec:kpls}, respectively.
Then, we summarize the two GEK approaches that already appeared in the literature in Sections~\ref{sec:ingek} and~\ref{sec:dgek}, followed by the development of the GE-KPLS approach Section~\ref{sec:GE-KPLS}.
We then compare the proposed GE-KPLS approach to the previously developed methods on analytic and engineering cases in Section~\ref{sec:result}.
Finally, we summarize our conclusions in Section~\ref{sec:concl} after presenting limitations of our approach in Section~\ref{sec:limitation}.

\section{Kriging surrogate modeling}

In this section we introduce the notation and briefly describe the theory behind kriging and KPLS.
The first step in the construction of surrogate models is the selection of sample points $\x^{(i)} \in \mathbb{R}^d$, for $i=1,\dots,n$, where $d$ is the number of inputs and $n$ is the number of sampling points.
We can denote this set of sample points as a matrix,
\begin{equation}
    \X = \left[{\x^{(1)}}^T,\ldots,{\x^{(n)}}^T\right]^T .
\end{equation}
Then, the function to be modeled is evaluated at each sample point.
We assume that the function to be modeled is deterministic, and we write it as $f:B\longrightarrow\mathbb{R}$, where, for simplicity, $B$ is a hypercube expressed by the product between the intervals of each direction space.
We obtain the outputs $\y=\left[y^{(1)},\ldots,y^{(n)}\right]^T$ by evaluating the function
\begin{equation}
\label{eq:evalFun}
y^{(i)}=f\left(\x^{(i)}\right),\hspace{6pt}\text{for} \, i =1,\dotsc,n .
\end{equation}
With the choice and evaluation of sample points we have $\left(\X,\y\right)$, which we can now use to construct the surrogate model.

\subsection{Conventional kriging}
\label{sec:krig}

\cite{Matheron1963} developed the theoretical basis of the kriging approach based on the work of~\cite{Krige1951}.
The kriging approach has since been then extended to the fields of computer simulation~\citep{Sacks1989a,Sacks1989b} and machine learning~\citep{Welch1992}.
The kriging model, also known as Gaussian process regression~\citep{Rasmussen2006}, is essentially an interpolation method.
The interpolated values are modeled by a Gaussian process with mean $\mu(.)$ governed by a prior spatial covariance function $k(.,.)$.
The covariance function $k$ can be written as
\begin{equation}
\label{eq:covFun}
k(\x,\x')=\sigma^2r(\x,\x')=\sigma^2r_{\x\x'},\hspace{6pt}\forall \x,\x'\in B,\\
\end{equation}
where $\sigma^2$ is the process variance and $r_{\x\x'}$ is the spatial correlation function between $\x$ and $\x'$.
The correlation function $r$ depends on hyperparameters $\theta$, which need be estimated.
In this paper, we use the Gaussian exponential correlation function for all the numerical results presented in Section~\ref{sec:result}.
\begin{equation}
\label{eq:GaussCorr}
r_{\x\x'}=\prod\limits_{i=1}^d\exp\left(-\theta_i\left(x_i-{x'}_i\right)^{2}\right), \quad \forall \theta_i\in \mathbb{R}^+,\,\forall \x,\x'\in B,
\end{equation}
Through this definition, the correlation between two points is related to the distance between the corresponding points $\x$ and $\x'$
This is a function that quantifies resemblance degree between any two points in the design space.

Let us now define the stochastic process $Y(x)=\mu +Z(x)$, where $\mu$ is an unknown constant, and $Z(x)$ is a realization of a stochastic Gaussian process with $Z\sim {\cal N}(0,\sigma^2)$.
In this study, we use the ordinary kriging model, where $\mu(x)=\mu=\text{constant}$.
To construct the kriging model, we need to estimate a set of unknown parameters: $\theta$, $\mu$, and $\sigma^2$.
To this end, we use the maximum likelihood estimation method.
In practice, we use the natural logarithm to simplify the likelihood maximization
\begin{equation}
\label{eq:logmle}
-\frac{1}{2}\big[n\ln( 2\pi\sigma^2)+\ln \boldsymbol(\det \R\boldsymbol)+(\y - \bf{1}\mu)^t\R^{-1}(\y-\bf{1}\mu)/\sigma^2\big],
\end{equation}
where \({\bf 1}\) denotes an \(n\)-vector of ones.

First, we assume that the hyperparameters $\theta$ are known, so $\mu$ and $\sigma^2$ are given by
\begin{equation}
\label{eq:betaOrd}
\hat{\mu}= \left({\bf 1}^T\R^{-1}{\bf 1}\right)^{-1}{\bf 1}^T\R^{-1}\y,
\end{equation}
where \({\bf R}=[\textbf{r}_{\x^{(1)}\X},\ldots,\textbf{r}_{\x^{(n)}\X}]\) is the correlation matrix with \(\textbf{r}_{\x \X}=[r_{\x\x^{(1)}},\ldots,r_{\x\x^{(n)}}]^T\) and
\begin{equation}
\label{eq:sigmaCarreOrd}
\hat{\sigma}^2=\frac{1}{n}\left(\y-{\bf 1}\hat{\mu}\right)^T{\bf R}^{-1}\left(\y-{\bf 1}\hat{\mu}\right).
\end{equation}
In fact, Equations~\eqref{eq:betaOrd} and~\eqref{eq:sigmaCarreOrd} are given by taking derivatives of the likelihood function and setting to zero.
Next, we insert both equations into the expression~\eqref{eq:logmle} and remove the constant terms, so the so-called concentrated likelihood function that depends only on $\theta$ is given by
\begin{equation}
\label{eq:concMLE}
-\frac{1}{2}\big[n\ln(\sigma^2(\theta))+\ln \boldsymbol(\det \R(\theta)\boldsymbol)\big],
\end{equation}
where $\sigma(\theta)$ and $\R(\theta)$ denote the dependency with $\theta$.
A detailed derivation of these equations is provided by~\cite{Forrester2008} and~\cite{Kleijnen2015}.
Finally, the best linear unbiased predictor, given the outputs \(\y\), is
\begin{equation}
\label{eq:predict}
\hat{y}(\x) = \hat{\mu} + \textbf{r}_{\x \X}^T\R^{-1}\left(\y-\textbf{1}\hat{\mu}\right),\hspace{6pt}\forall \x\in B.
\end{equation}

Since there is no analytical solution for estimating the hyperparameters $\theta$, it is necessary to use numerical optimization to find the hyperparameters $\theta$ that maximize the likelihood function.
This step is the most challenging in the construction of the kriging model.
This is because, as previously mentioned, this estimation involves maximizing the likelihood function, which is often multimodal~\citep{Mardia1989}.
Maximizing this function becomes prohibitive for high-dimensional problems ($d>10$) due to the cost of computing the determinant of the correlation matrix and the high number of evaluation needed for optimizing a high-dimensional multimodal problem.
This is the main motivation for the development of the KPLS approach, which we describe next.

\subsection{KPLS(K)---Accelerating kriging construction with partial-least squares regression}
\label{sec:kpls}

As previously mentioned, the estimation of the kriging hyperparameters can be time consuming, particularly for high-dimensional problems.
\citet{Bouhlel2016a} recently developed an approach that reduces the computational cost while maintaining accuracy by using the PLS regression during the hyperparameters estimation process.
PLS regression is a well-known method for handling high-dimensional problems, and consists in maximizing the variance between input and output variables in a smaller subspace, formed by principal components---or latent variables.
PLS finds a linear regression model by projecting the predicted variables and the observable variables to a new space.
The elements of the principal direction, that is a vector defining the direction of the associated principal component, represent the influence of each input on the output.
On the other hand, the hyperparameters $\theta$ represent the range in any direction of the space.
Assuming, for instance, that certain values are less significant in the $i^\text{th}$ direction, the corresponding $\theta_i$ should have a small value.
Thus, the key idea behind the construction of the KPLS model is the use of PLS information to adjust hyperparameters of the kriging model.

We compute the first principal component \({\bf t}_1\) by seeking the direction \(\w^{(1)}\) that maximizes the squared covariance between \({\bf t}_1 = \X\w^{(1)}\) and \(\y\), i.e.,
\begin{equation}
\label{eq:firstComp}
\w^{(1)} =
\left\lbrace \begin{aligned}
\arg\max\limits_{\w} &\,\,\,\w^T \X^T \y \y^T \X \w \\
\,\text{ such that}&\,\,\,\w^T \w\,=\,1.
\end{aligned} \right.
\end{equation}
Next, we compute the residual matrix from \(\X^{(0)}\stackrel{}{\leftarrow}\X\) space and from \(\y^{(0)}\stackrel{}{\leftarrow}\y\) using
\begin{equation}
\label{eq:sys11}
\begin{aligned}
\X^{(1)} &= \X^{(0)} - {\bf t}_1 \p^{(1)},\\
\y^{(1)} &= \y^{(0)} - c_1 {\bf t}_1,\\
\end{aligned}
\end{equation}
where \(\p^{(1)}\) (a \(1\times d\) vector) contains the regression coefficients of the local regression of \(\X\) onto the first principal component \({\bf t}_1\) (an \(n\times 1\) vector), and \(c_1\) is the regression coefficient of the local regression of \(\y\) onto the first principal component \({\bf t}_1\).
Next, the second principal component---orthogonal to the first principal component---can be sequentially computed by replacing \(\X^{(0)}\) by \(\X^{(1)}\) and  \(\y^{(0)}\) by \(\y^{(1)}\) to solve the maximization problem~\eqref{eq:firstComp}.
The same approach is used to iteratively compute the other principal components.

The computed principal components represent the  new coordinate system obtained upon rotating the original system with axes, \(x_1,\dotsc,x_d\)~\citep{Alberto2012}.
The $l^\text{th}$ principal component \({\bf t}_l\) is
\begin{equation}
{\bf t}_l = \X^{(l-1)}\w^{(l)} = \X\w_*^{(l)}, \quad \text{for}\, l=1,\dotsc,h.
\end{equation}
The matrix \(\W_*=\left[\w_*^{(1)},\dotsc,.\w_*^{(h)}\right]\) is obtained by using the following formula~\citep[pg.~114]{Tenenhaus1998}
\begin{equation}
\W_*=\W\left({\bf P}^T\W\right)^{-1},
\end{equation}
where \(\W=\left[\w^{(1)},\dots,\w^{(h)}\right]\) and \({\bf P}=\left[{\p^{(1)}}^T,\dots,{\p^{(h)}}^T\right]\).
If $h=d$, the matrix \(\W_*=\left[\w_*^{(1)},\dots,\w_*^{(d)}\right]\) rotates the coordinate space \((x_1,\dotsc,x_d)\) to the new coordinate space \((t_1,\dots,t_d)\), which follows the principal directions \(\w^{(1)},\dots,\w^{(d)}\).
More details on the PLS method can be found in the literature~\citep{Helland1988,Frank1993,Alberto2012}.

The PLS method gives information on any variable contribution to the output.
Herein lies the idea developed by~\cite{Bouhlel2016a}, which consists in using information provided by PLS to add weights on the hyperparameters $\theta$.
For $l=1,\dotsc,h$, the scalars \(w^{(l)}_{*1},\dotsc,w^{(l)}_{*d}\) are interpreted as measuring the importance of \(x_1,\dotsc,x_d\), respectively, for constructing the $l^\text{th}$ principal component where its correlation with the output \(y\) is maximized.

To construct the KPLS kernel, we first define the linear map $F_l$ by
\begin{equation}
\label{eq:fl}
\begin{aligned}
F_l:&B &\longrightarrow&B\\
&\x& \longmapsto &\left[w^{(l)}_{*1} x_1,\dots,w^{(l)}_{*d} x_{d}\right]\hspace{-0.13cm},
\end{aligned}
\end{equation}
for $l=1,\dotsc,h$.
By using the mathematical property that the tensor product of several kernels is a kernel, we build the KPLS kernel
\begin{equation}
\label{eq:noyau1h}
k_{1:h}(\x,\x') = \prod\limits_{l=1}^h k_l\bm( F_l\left(\x\right),F_l\left({\x'}\right)\bm),\,\forall \x,\x'\in B,
\end{equation}
where \(k_l:B \times B \rightarrow \mathbb{R}\) is an isotropic stationary kernel, which is invariant when translated.
More details of this construction are described by~\cite{Bouhlel2016a}.

If we use the Gaussian correlation function~\eqref{eq:GaussCorr} and in this construction~\eqref{eq:noyau1h}, we obtain
\begin{equation}
\label{eq:GaussCorrKPLS}
k(\x,\x')=\sigma^2\displaystyle{\prod\limits_{l=1}^h\prod\limits_{i=1}^d}\exp\left[-\theta_l
\left(w^{(l)}_{*i}x_i-w^{(l)}_{*i}{x'}_i\right)^{2}\right],\forall\ \theta_l\in[0,+\infty[,\hspace{6pt}\forall \x,\x'\in B.
\end{equation}
The KPLS method reduces the number of hyperparameters to be estimated from $d$ to $h$, where $h<<d$, thus drastically decreasing the time to construct the model.

\citep{Bouhlel2016b} proposed another method to construct a KPLS-based model for high-dimensional problems, the so-called KPLSK.
This method is applicable only when covariance functions used by KPLS are of the exponential type (e.g., all Gaussian), then the covariance function used by KPLSK is exponential with the same form as the KPLS covariance.
This method is basically a two-step approach for optimizing the hyperparameters.
The first step consists in optimizing the hyperparameters of a KPLS covariance, this is by using a gradient-free method on $h$ hyperparameters for a global optimization in the reduced space.
The second step consists in optimizing the hyperparameters of a conventional kriging model by using a gradient-based method and the solution of the first step, this is for a local improvement of the solution provided by the first step into the original space ($d$ hyperparameters).
The idea of this approach is to use a gradient-based method, which is more efficient than a gradient-free method, with an initial guess for the construction of a conventional kriging model.

The solution of the first step with $h$ hyperparameters is expressed in the bigger space with $d$ hyperparameters using a change of variables.
By using Equation~\eqref{eq:GaussCorrKPLS} and the change of variable $\eta_i = \sum\limits_{l=1}^h\theta_l{w_{*i}^{(l)}}^2$, we get
\begin{equation}
\begin{array}{lll}
\sigma^2\prod\limits_{l=1}^h\prod\limits_{i=1}^d\exp{\left(-\theta_l{w_{*i}^{(l)}}^2(x_i-{x'}_i)^2\right)} &=&\sigma^2\exp\left(\sum\limits_{i=1}^d\sum\limits_{l=1}^h-\theta_l{w_{*i}^{(l)}}^2(x_i-{x'}_i)^2\right)\\
&=&\sigma^2\exp\left(\sum\limits_{i=1}^d-\eta_i(x_i-{x'}_i)^2\right)\\
&=&\sigma^2\prod\limits_{i=1}^d\exp\left(-\eta_i(x_i-{x'}_i)^2\right).\\
\end{array}
\end{equation}
This is the definition of a Gaussian kernel given by Equation~\eqref{eq:GaussCorr}.
Therefore, each component of the starting point for the gradient-based optimization uses a linear combination of the hyperparameters' solutions from the reduced space.
This allows the use of an initial line search along a hypercube of the original space in order to find a relevant starting point.
Furthermore, the final value of the likelihood function (KPLSK) is improved compared to the one provided by KPLS.
The KPLSK model is computationally more efficient than a kriging model and slightly more costly than KPLS.

\section{Gradient-enhanced kriging}
\label{sec:gek}

If the gradient of the output function at the sampling points is available, we can use this information to increase the accuracy of the surrogate model.
Since a gradient consists of $d$ derivatives, adding this much information to the function value at each sampling point has the potential to enrich the model immensely.
Furthermore, when the gradient is computed using an adjoint method, whose computational cost is similar that of a single function evaluation and independent of $d$, this enrichment can be obtained at much lower computational cost than evaluating $d$ new function values.

Various approaches have been developed for GEK, and two main formulations exist: indirect and direct GEK.
In the following, we start with a brief review of these formulations, and then we present GE-KPLS---our novel approach.

\subsection{Indirect gradient-enhanced kriging}
\label{sec:ingek}

The indirect GEK method consists in using the gradient information to generate new points around the sampling points via linear extrapolation.
In each direction of each sampling point, we add one point by computing the FOTA
\begin{equation}
\label{eq:oft}
y\left(\x^{(i)}+\Delta x_{j}\e^{(j)}\right) = y\left(\x^{(i)}\right)+\frac{\partial y\left(\x^{(i)}\right)}{\partial x_j}\Delta x_{j},
\end{equation}
where $i=1,\dotsc,n$, $j=1,\dotsc,d$, $\Delta x_{j}$ is the step added in the $j^{\text{th}}$ direction, and $\e^{(j)}$ is the $j^\text{th}$ row of the $d \times d$ identity matrix.
The indirect GEK method does not require a modification of the kriging code.
However, the resulting correlation matrix can rapidly become ill-conditioned, since the columns of the matrix due to the FOTA are almost collinear.
Moreover, this method increases the size of the correlation matrix from $n\times n$ to $n(d+1) \times n(d+1)$.
Thus, the computational cost to build the model becomes prohibitive for high-dimensional problems.

\subsection{Direct gradient-enhanced kriging}
\label{sec:dgek}

In the direct GEK method, derivative values are included in the vector $\y$ from Equation~\eqref{eq:predict}.
This vector is now
\begin{equation}
\y =
\begin{bmatrix}
y\left(\x^{(1)}\right), \dots,y\left(\x^{(n)}\right),\frac{\partial y\left(\x^{(1)}\right)}{\partial x_1},\dots,\frac{\partial y\left(\x^{(1)}\right)}{\partial x_d},\dots,\frac{\partial y\left(\x^{(n)}\right)}{\partial x_d}
\end{bmatrix}^T,
\label{eq:geky}
\end{equation}
with a size of $n(d+1) \times 1$.
The vector of ones from Equation~\eqref{eq:predict} also has the same size and is
\begin{equation}
\textbf{1}= [
\overbrace{
1,\dots,1
}^{n},\overbrace{0,\dots,0}^{nd}]^T .
\label{eq:gek1}
\end{equation}

The size of the correlation matrix increases to $n(d+1) \times n(d+1)$, and contains four blocks that include the correlation between the data and themselves, between  the gradients and themselves, between the data and gradients, and between the gradients and data.
Denoting the GEK correlation matrix by $\stackrel{.}{\R}$, we can write
\begin{equation}
\stackrel{.}{\R}=
\begin{bmatrix}
r_{\x^{(1)}\x^{(1)}}&\dots&r_{\x^{(1)}\x^{(n)}}&\frac{\partial r_{\x^{(1)}\x^{(1)}}}{\partial \x^{(1)}}&\dots&\frac{\partial r_{\x^{(1)}\x^{(n)}}}{\partial \x^{(n)}}\\
\vdots&\ddots&\vdots&\vdots&\ddots&\vdots\\
r_{\x^{(n)}\x^{(1)}}&\dots&r_{\x^{(n)}\x^{(n)}}&\frac{\partial r_{\x^{(n)}\x^{(1)}}}{\partial \x^{(1)}}&\dots&\frac{\partial r_{\x^{(n)}\x^{(n)}}}{\partial \x^{(n)}}\\
\frac{\partial r_{\x^{(1)}\x^{(1)}}}{\partial \x^{(1)}}^T&\dots&\frac{\partial r_{\x^{(1)}\x^{(n)}}}{\partial \x^{(1)}}^T&\frac{\partial^2 r_{\x^{(1)}\x^{(1)}}}{\partial^2 \x^{(1)}}&\dots&\frac{\partial^2 r_{\x^{(1)}\x^{(n)}}}{\partial \x^{(1)}\partial \x^{(n)}}\\
\vdots&\ddots&\vdots&\vdots&\ddots&\vdots\\
\frac{\partial r_{\x^{(n)}\x^{(1)}}}{\partial \x^{(n)}}^T&\dots&\frac{\partial r_{\x^{(n)}\x^{(n)}}}{\partial \x^{(n)}}^T&\frac{\partial^2 r_{\x^{(n)}\x^{(1)}}}{\partial \x^{(n)}\partial \x^{(1)}}&\dots&\frac{\partial^2 r_{\x^{(n)}\x^{(n)}}}{\partial^2 \x^{(n)}}
\end{bmatrix},
\label{eq:corgek}
\end{equation}
where, for $i,j=1,\dots,n$, $\partial r_{\x^{(i)}\x^{(j)}}/\partial \x^{(i)}$, $\partial r_{\x^{(i)}\x^{(j)}}/\partial \x^{(j)}$ and $\partial^2 r_{\x^{(i)}\x^{(j)}}/\partial \x^{(i)}\partial \x^{(j)}$ are given by
\begin{equation}
\frac{\partial r_{\x^{(i)}\x^{(j)}}}{\partial \x^{(i)}} =
\begin{bmatrix}
\frac{\partial r_{\x^{(i)}\x^{(j)}}}{\partial x^{(i)}_k}=-2\theta_k \left(x^{(i)}_k-x^{(j)}_k\right)r_{\x^{(i)}\x^{(j)}}
\end{bmatrix}_{k=1,\dots,d},
\end{equation}
\begin{equation}
\frac{\partial r_{\x^{(i)}\x^{(j)}}}{\partial \x^{(j)}} =
\begin{bmatrix}
\frac{\partial r_{\x^{(i)}\x^{(j)}}}{\partial x^{(j)}_k}=2\theta_k \left(x^{(i)}_k-x^{(j)}_k\right)r_{\x^{(i)}\x^{(j)}}
\end{bmatrix}_{k=1,\dots,d},
\end{equation}
\begin{equation}
\frac{\partial^2 r_{\x^{(i)}\x^{(j)}}}{\partial \x^{(i)}\partial \x^{(j)}} =
\begin{bmatrix}
\frac{\partial^2 r_{\x^{(i)}\x^{(j)}}}{\partial x^{(i)}_k\partial x^{(j)}_l}=-4\theta_k\theta_l \left(x^{(i)}_k-x^{(j)}_k\right)\left(x^{(i)}_l-x^{(j)}_l\right)r_{\x^{(i)}\x^{(j)}}
\end{bmatrix}_{k,l=1,\dots,d}.
\end{equation}
Once the hyperparameters $\theta$ are estimated, the GEK predictor for any untried $\x$ is given by
\begin{equation}
\label{eq:predictGEK}
\hat{y}(\x) = \hat{\mu} + \stackrel{.}{\textbf{r}}^T_{\x \X}\stackrel{.}{\R}^{-1}\left(\y-\textbf{1}\hat{\mu}\right),\hspace{6pt}\forall \x\in B,
\end{equation}
where the correlation vector contains correlation values of an untried point $\x$ to each training point from $\X=\left[\x^{(1)},\dotsc,\x^{(n)}\right]$ and is
\begin{equation}
\stackrel{.}{\bf{r}}_{\x \X}=
\begin{bmatrix}
r_{\x\x^{(1)}}&\dots&r_{\x\x^{(n)}}&\frac{\partial r_{\x^{(1)}\x}}{\partial \x^{(1)}}&\dots&\frac{\partial r_{\x^{(n)}\x}}{\partial \x^{(n)}}
\end{bmatrix}^T.
\end{equation}

Unfortunately, the correlation matrix $\stackrel{.}{\R}$ is dense, and its size increases quadratically both with the number of variables $d$ and the number of samples $n$.
In addition, $\stackrel{.}{\R}$ is not symmetric, which makes it more costly to invert.
In the next section, we develop a new approach that uses the gradient information with a controlled increase in the size of the correlation matrix $\stackrel{.}{\R}$.

\subsection{GE-KPLS---Gradient-enhanced kriging with partial-least squares method}
\label{sec:GE-KPLS}

While effective in several problems, GEK methods are still vulnerable to a number of weaknesses.
As previously discussed, the weaknesses have to do with the rapid growth in the size of the correlation matrix when the number of sampling points, the number of inputs, or both, become large.
Moreover, high-dimensional problems lead to a high number of hyperparameters to be estimated, and this results in challenging problems in the maximization of the likelihood function.
To address these issues, we propose the GE-KPLS approach, which exploits the gradient information with a slight increase of the size of the correlation matrix but reduces the number of hyperparameters.

\subsubsection{Model construction}

\begin{figure}[H]
\centering
\subfloat[Box--Behnken design.\label{BoxBehnken}]{\includegraphics[scale=0.38]{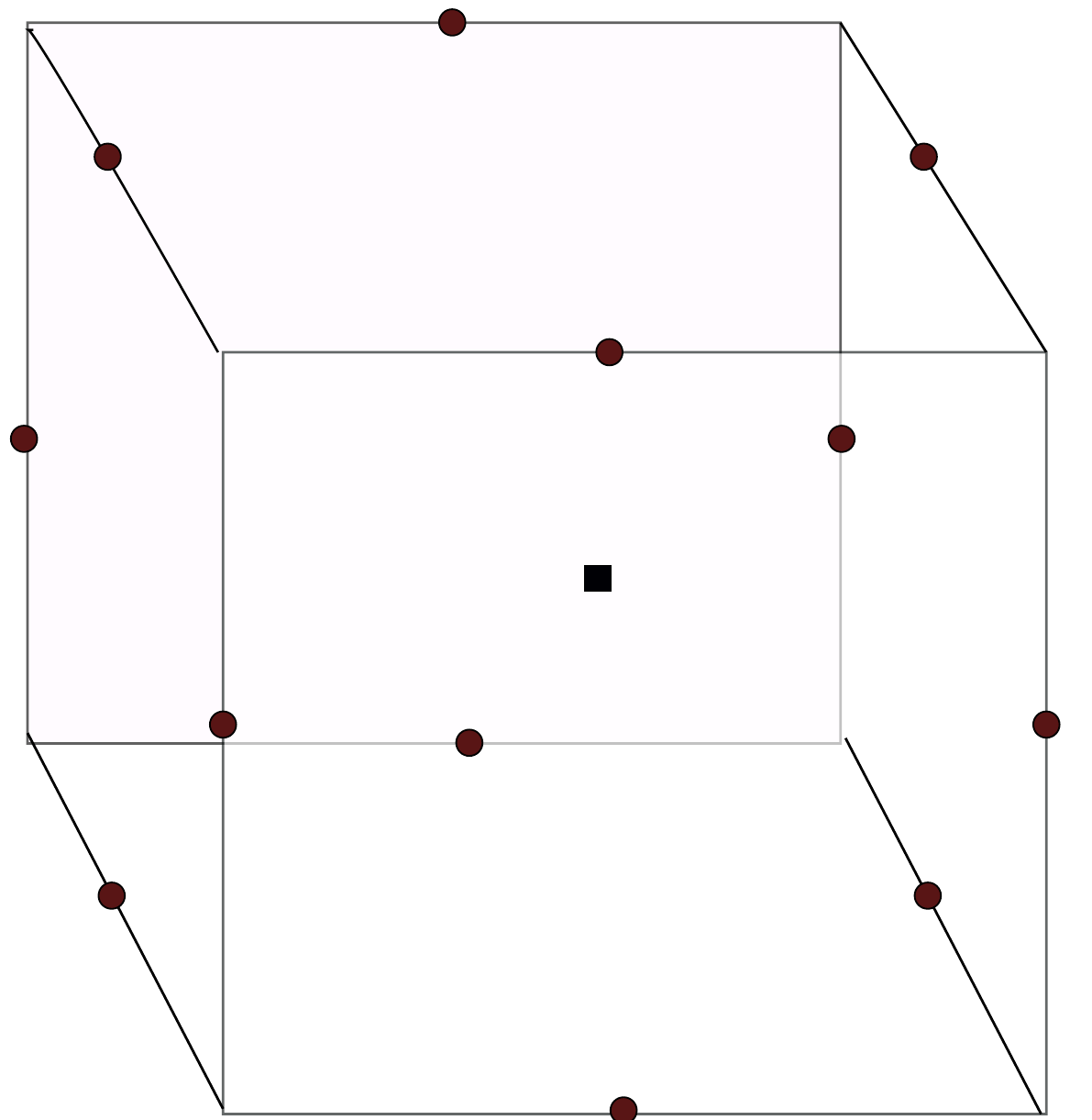}}\hspace{5em}
\subfloat[Forward and backward variations of each direction.\label{plan}]{\includegraphics[scale=0.38]{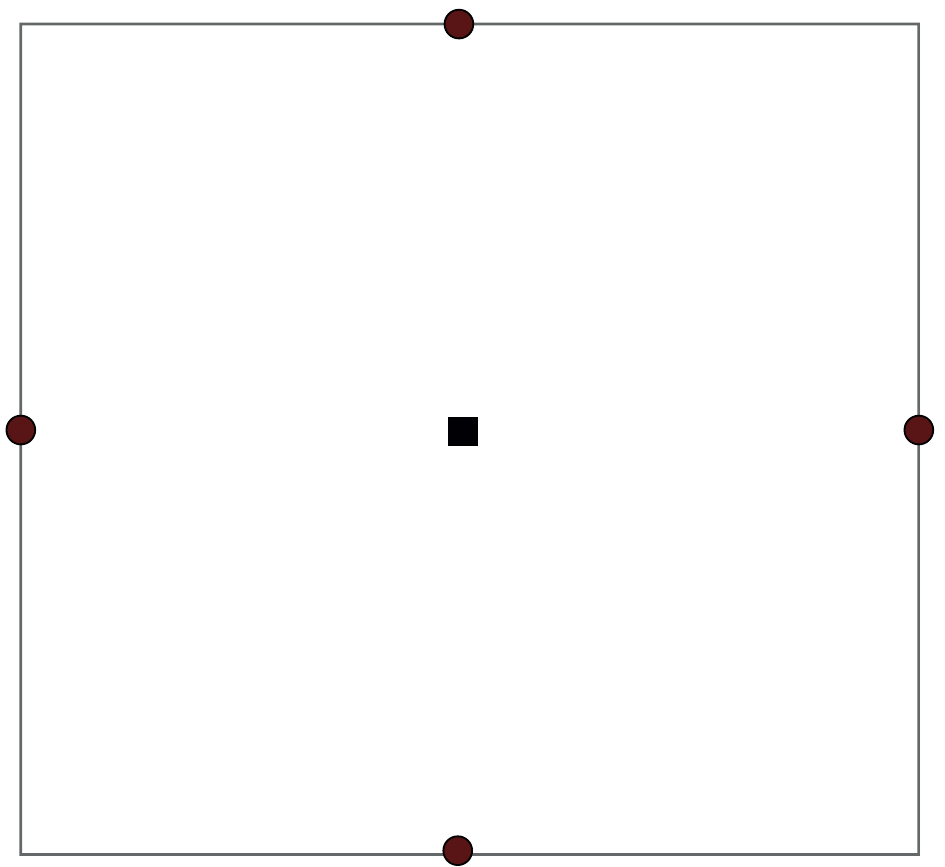}}\\
\caption{\label{box_design}The circular and rectangular points are the new generated points and sampling points, respectively.}
\end{figure}

The key idea of the proposed method consists in using the PLS method around each sampling point; we apply the PLS method several times, each time on a different sampling point.
To this end, we use the FOTA~\eqref{eq:oft}, to generate a set of points around each sampling point.
These new approximated points are constructed either by a Box--Behnken design~\citep[Ch. 11, Sec. 6]{Box2005} when $d \geq 3$ (Figure~\ref{BoxBehnken}) or by a forward and backward variations in the $2d$-space (Figure~\ref{plan}).

PLS is applied to GEK as follows.
Suppose we have a sets of points $\mathcal{S}=\{\mathcal{S}_i,\,\forall i=1,\dotsc,n\}$, where each set of points is defined by the sampling point $\left(\x^{(i)},y^{(i)}\right)$ and the set of approximating points generated by FOTA on the Box--Behnken design when $d\geq 3$, or on forward and backward variations in the $2d$-space.
We then apply the PLS method on each set of points $\mathcal{S}_i$ to get the local influence of each direction space.
Next, we compute the mean of the $n$ coefficients $\left|\w^{(l)}_*\right|$ for each principal component $l=1,\dotsc,h$.
Denoting these new coefficients by $\w^{(l)}_{\text{av}}$, we replace the Equation~\eqref{eq:fl} by
\begin{equation}
\label{eq:flnew}
\begin{aligned}
F_l:&B &\longrightarrow&B\\
&\x& \longmapsto &\left[w^{(l)}_{\text{av}_1} x_1,\dots,w^{(l)}_{\text{av}_d} x_{d}\right] .
\end{aligned}
\end{equation}
Finally, we follow the same construction used for the KPLS model by substituting $\w^{(l)}_*$ by $\w^{(l)}_{\text{av}}$.
Thus Equation~\eqref{eq:GaussCorrKPLS} becomes
\begin{equation}
\label{eq:GaussCorrGE-KPLS}
k(\x,\x')=\sigma^2\displaystyle{\prod\limits_{l=1}^h\prod\limits_{i=1}^d}\exp\left[-\theta_l
\left(w^{(l)}_{\text{av}_i}x_i-w^{(l)}_{\text{av}_i}{x'}_i\right)^{2}\right], \quad \forall\ \theta_l\in[0,+\infty[,\quad \forall \x,\x'\in B.
\end{equation}

In the next section, we describe how we control the size of the correlation matrix to obtain the best trade-off between the accuracy of the model and the computational time.

\subsubsection{Controlling the size of the correlation matrix}

We have seen in Section~\ref{sec:ingek} that the construction of the indirect GEK model consists in adding $d$ points around each sampling points.
Since the size of the correlation matrix is $n(d+1)\times n(d+1)$, this leads to a dramatic increase in the matrix size.
In addition, the added sampling points are close to each other, leading to an ill-conditioned correlation matrix.
Thus, the inversion of the correlation matrix becomes difficult and computationally prohibitive for large numbers of sampling points.
However, adding only relevant points improves both the correlation matrix condition number and the accuracy of the model.

In the previous section, we locally apply the PLS technique with respect to each sampling point, which provides the influence of each input variable around that point.
The idea here is to add only $m$ approximating points ($m\in[1,d]$) around each sampling point, where $m$ is the corresponded highest coefficients of PLS.
To this end, we consider only coefficients given by the first principal component, which usually contains the most useful information.
Using this construction, we improve the accuracy of the model with respect to relevant directions and increase the size of the correlation matrix to only $n(m+1)\times n(m+1)$, where $m << d$.

Algorithm~\ref{alg.GE-KPLS} describes how the information flows through the construction of the GE-KPLS model from sampling data to the final predictor.
Once the training points with the associated derivatives, the number of principal components and the number of extra points are initialized, we compute $\w^{(1)}_\text{av},\dotsc,\w^{(h)}_\text{av}$.
To this end, we construct $\mathcal S_i$, apply the PLS on $\mathcal S_i$, and select the $m^\text{th}$ most influential Cartesian directions from the first principal component, this is for each sample point.
Then, we maximize the concentrated likelihood function given by Equation~\eqref{eq:concMLE}, and finally, we express the prediction $\hat{y}$ given by Equation~\eqref{eq:predict}.

\begin{algorithm}
\caption{Construct GE-KPLS model}
\SetKwInOut{Input}{input}\SetKwInOut{Output}{output}
\Input{\(\left(\X,\y,\frac{\partial \y}{\partial \X},h,m\right)\)}
\Output{$\hat{y}(\x)$}
\For{$i \leq n$}{
$\mathcal S_i$\tcp*{To generate a set of approximating points}
$\mathcal S_i\stackrel{\text{PLS}}{\longrightarrow} \left(\w^{(1)}_*,\dotsc, \w^{(h)}_*\right)$
$\max \left|\w^{(1)}_*\right|$\tcp*{To select the $m^\text{th}$ most influential coefficients}
}
$\w^{(1)}_\text{av},\dotsc,\w^{(h)}_\text{av}$\tcp*{To compute the average of the PLS coefficients}
$\theta_1,\dotsc,\theta_h$\tcp*{To estimate the hyperparameters}
$\hat{y}(\x)$
\label{alg.GE-KPLS}
\end{algorithm}

In the GE-KPLS method, the PLS technique is locally applied around each sampling point instead of the whole space, as in the KPLS model.
This enables us to identify the locally influence of the input variables where sampling points are located.
By taking the mean of all the local input variable influences, we expect to obtain a good estimation of the global input variable influences.
The main computational advantages in such construction are the reduced number of hyperparameters to be estimated---since $h << d$---and the reduced size of the correlation matrix---$n(m+1) \times n(m+1)$, with $m << d$, compared to $n(d+1) \times n(d+1)$ for the conventional indirect and direct GEK models.

In the next section, our proposed methods are performed on high-dimensional benchmark functions and engineering cases.

\section{Numerical experiments}
\label{sec:result}

To evaluate the computational cost and accuracy of the proposed GE-KPLS method, we perform a series of numerical experiments where we compare GE-KPLS to other previously developed models for a number of benchmark functions.
The first set of functions consists of two different analytic functions of varying dimensionality given by
\begin{equation}
\label{eq:y1}
y_1(\x) = \sum\limits_{i=1}^dx_i^2,\hspace{1cm}-10\leq x_i\leq 10,\text{ for }i=1,\ldots,d.
\end{equation}
\begin{equation}
\label{eq:y2}
y_2(\x) = x_1^3+\sum\limits_{i=2}^{d}x_i^2,\hspace{1cm}-10\leq x_i\leq 10,\text{ for }i=1,\ldots,d.
\end{equation}
The second set of functions is a series of eight functions corresponding to engineering problems listed in Table~\ref{dimSampEng}.
\begin{table}[H]
\caption{\label{dimSampEng}Definition of engineering functions.}
\centering
\begin{tabular}{@{}rlcccl@{}}
\toprule
&Problem& $d$ & $n_1$& $n_2$ & Reference \\
\midrule
P$_1$ &  Welded beam & $2$ & 10&20&\citet{Deb1998} \\
P$_2$& Welded beam &2&10&20&\citet{Deb1998} \\
P$_3$& Welded beam &4&20&40&\citet{Deb1998} \\
P$_4$& Borehole &8&16&80&\citet{Morris1993} \\
P$_5$& Robot &8&16&80&\citet{An2001} \\
P$_6$& Aircraft wing &10&20&100&\citet{Forrester2008} \\
P$_7$& Vibration &15&75&150&\citet{Wang2006} \\
P$_8$& Vibration &15&75&150&\citet{Wang2006} \\
\bottomrule
\end{tabular}
\end{table}

Since the GEK model does not perform well, especially when the number of sampling points is relatively high as discussed previously, we performed three different studies.
The first study consists in comparing GE-KPLS with the indirect GEK and ordinary kriging models on the two analytic functions defined by Equations \eqref{eq:y1} and \eqref{eq:y2}.
The second and third studies, which use the same analytic functions as the first study and the engineering functions respectively, consist in comparing GE-KPLS with the ordinary kriging, KPLS and KPLSK models with an increased number of sampling compared to the first study.

The kriging, GEK, and KPLS(K) models, using the Gaussian kernels~\eqref{eq:GaussCorr} and~\eqref{eq:GaussCorrKPLS}, respectively, provide the benchmark results that we compared to the GE-KPLS model, using the Gaussian kernel~\eqref{eq:GaussCorrGE-KPLS}.
We use an unknown constant, $\mu$, as a trend for all model.
For the kriging experiments, we use the scikit-learn implementation~\citep{Pedregosa2011}.
Moreover, the indirect GEK method does not require a modification of the kriging source code, so we use the same scikit-learn implementation.

We vary the number of extra points, $m$, from 0 to 5 in all cases except for the first study, where in addition we use $m=d$, and also for the third study when the number of inputs is less than 5 input variables; e.g., $P_1$ from the engineering functions where $m=1$ and $m=d=2$.
To refer to the number of extra points, we denote GE-KPLS with $m$ extra points by GE-KPLS$m$.
We ran prior tests varying the number of principal components from 1 to 3 for the KPLS and KPLSK models, and 3 principal components always provided the best results.
Using more principal components than 3 becomes more costly and results in a very slight difference in terms of accuracy (more or less accurate depending on the problem).
For the sake of simplicity, we consider only results with 3 principal components for KPLS and KPLSK.
Similarly, the GE-KPLS method uses only 1 principal component, which was found to be optimal.

Because the GEK and GE-KPLS models use additional information (the gradient components) compared to other models and to make the comparison as fair as possible, the number of sampling points $n$ used to construct the GEK and GE-KPLS models is always twice less than the number of samples used for other models in all test cases.
This factor of two is to account for the cost of computing the gradient; when an adjoint method is available, this cost is roughly the same or less than the cost of computing the function itself~\cite{Martins2013a,Kenway2014a}.

To generate approximation points with FOTA, \citet{Laurenceau2008} recommend to use a step of $10^{-4}l_i$, where $l_i$ is the length between the upper and lower bounds in the $i^\text{th}$ direction.
However, we found in our test cases that the best step is not always $10^{-4}l_i$, so we performed an analysis to compute the best step value for the second and third studies.
The computational time needed to find the best step is not considered and we only report the computational time needed to construct the GE-KPLS models using this best step.
Because the GEK model is very expensive in some cases (see Section~\ref{sec:study1}), we only use the recommended step by \citet{Laurenceau2008} to perform the GE-KPLS and GEK methods for the first study.

To compare the accuracy of the various surrogate models, we compute the relative error (RE) for $n_v$ validation points as follows
\begin{equation}
\label{eq:RE}
\text{RE} = \frac{\lVert \bf{y}-\hat{\bf{y}}\rVert}{\lVert \bf{y}\rVert},
\end{equation}
where \(\hat{\bf{y}}\) is the surrogate model values evaluated at validation points, \(\bf{y}\) is the corresponding reference function values, and \(\lVert . \rVert\) is the $L_2$ norm.
Since in this paper we use explicit functions, the reference values can be assumed to have a machine epsilon of ${\cal O}\left(10^{-16}\right)$.
In addition, the function computations are fast, so generating a large set of random validation points is tractable.
We use $n_v=5,000$ validation points for all cases.
The sampling points and validation points are generated using the Latin hypercube implementation in the pyDOE toolbox~\citep{Abraham2009} using a \textit{maximin} and \textit{random} criteria, respectively.
We perform 10 trials for each case and we always plot the mean of our results.
Finally, all computations are performed on an Intel\textsuperscript{\textregistered} Core\textsuperscript{TM} i7-6700K 4.00\,GHz CPU .

\subsection{Numerical results for the analytical functions: first study}
\label{sec:study1}

To benchmark the performance of our approach, we first use the two analytical functions~\eqref{eq:y1} and~\eqref{eq:y2} and compare GE-KPLS$m$, for $m=1,\dots,5$ and $m=d$, to the GEK and kriging models.
For this study, we have added the case where $m=d$ (compared to the next two studies) to figure out the usefulness of the PLS method in our approach, since the number of extra points is the same as for the GEK model.
We vary the number of inputs for both functions from $d=20$ to $d=100$ by adding 20 dimensions at a time.
In addition, we vary the number of sampling points in all cases from $n=10$ to $n=100$ by adding 10 samples at the time for the GEK and GE-KPLS models.
For the construction of the kriging model, we use $2n$ sampling points for each case.

\begin{figure}[H]
\centering
\includegraphics[width=\linewidth]{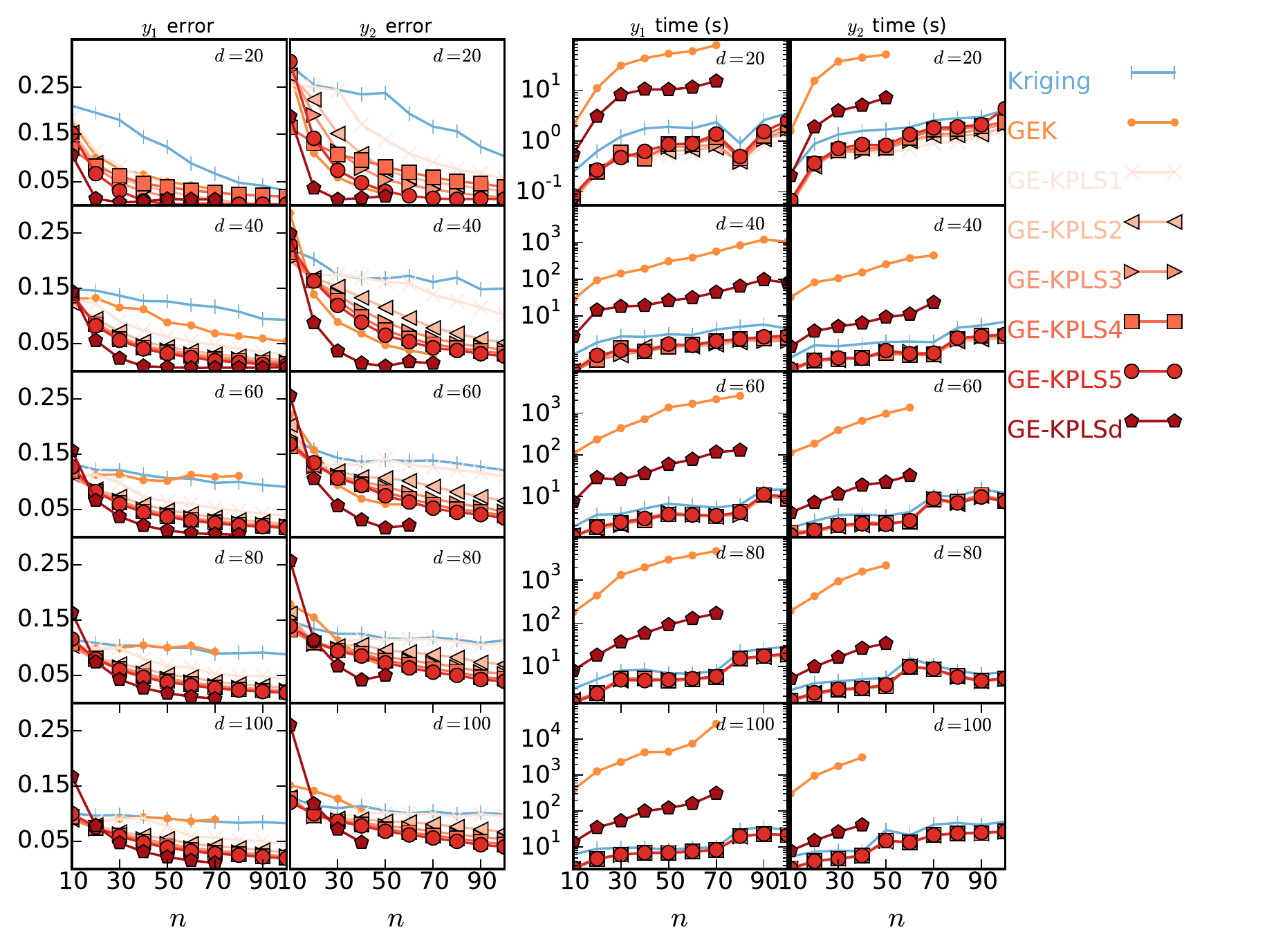}
\caption{\label{res:study1}Summary of the mean results for kriging, GEK, and GE-KPLS models applied to the analytical problems, based on 10 trials for each case.
The models are color coded according to the legend (upper right).}
\label{fig:analytic}
\end{figure}

Figure~\ref{res:study1} summarizes the results of this first study.
The first two columns show the RE for $y_1$ and $y_2$, respectively, and the other two columns show the computational time for the same two functions.
Each row shows the results  for increasing dimensionality, starting with $d=20$ at the top and ending at $d=100$ at the bottom.
The models are color coded as shown in the legend on the upper right.
In some cases, we could not reach 100 sampling points because of the ill-conditioned covariance matrix provided by the GEK model, which explains the missed experiments in all cases except for the $y_1$ function with $d=40$.

The GE-KPLS$d$ and GEK use the same points (training and approximating points) into their correlation matrices, and the difference between both models consists in reducing the number of hyperparameters by PLS for only the first model, so we can verify the scalability of the GE-KPLS model with the inputs variables (through the hyperparameters).
The GE-KPLS$d$ model yields a more accurate model compared to GEK in all cases except for $y_1$ when $d\geq 40$ and $n=10$, and for $y_2$ when $d > 40$ and $n=10$.
These results show the effectiveness of the PLS method in reducing the computational time especially when $d>60$.
For example, the computational time for the $y_2$ function with $d=80$ and $n\leq 50$ is less than 45\,s for PLS compared to the computational time of GEK where it reaches 42\,min.
Therefore, the PLS method improves the accuracy of the model \emph{and} reduces the computational time needed to construct the model.

Even though the RE-convergence of GE-KPLS$d$ is the most accurate, the GE-KPLS$m$ models for $m=1,\dots,5$ are in some cases preferable.
For example, the GE-KPLS$m$ models for $m=1,\dots,5$ are over 37 times faster than the GE-KPLS$d$ model with about a 1.5\% of lost in term of error for $y_1$ with $d=100$ and $n=70$.
In addition, including $d$ extra points around each sampling point leads to ill-conditioned matrices, in particular when the number of sampling points is high.
Furthermore, the GE-KPLS$m$ models for $m=1,\dots,5$ always yield a lower RE and decreased computational time compared to kriging.
When comparing kriging and GE-KPLS$m$ for $m=1,\dots,5$, the computational time of GE-KPLS$m$ is 10\,s lower for all cases and the RE is 10\% better in some cases; e.g. the $y_1$ function with $d=20$, $n=30$ using a GE-KPLS5 model.
Compared to GEK, GE-KPLS$m$ for $m=1,\dots,5$ has a better RE convergence with the $y_1$ function, and the RE convergence on $y_2$ is slightly better with GEK when $d \le 80$.
In addition, GE-KPLS$m$ has lower computational times compared to GEK; e.g. the time needed to construct GE-KPLS$m$ with $m=1,\dotsc,5$ for $y_1$ with 100$d$ and 70 points is between 7\,s and 9\,s compared to about 27000\,s for GEK.

We also note that the difference between the two functions $y_1$ and $y_2$ is only about the first term $x_1$, and despite these similarities, the results for both functions are different.
For example, the RE-convergence of all GE-KPLS$m$ are better than the GEK convergence for $y_1$ with $d=40$, which is not the case for $y_2$ with $d=40$.
Therefore, it is safer to make a new selection of the best model for a function even though we know the best model for a similar function to the first one.

Finally, the construction of the GEK model could be prohibitive in terms of computational time.
For instance, we need about 7.4 hours to construct a GEK model for $y_1$ with $d=100$ and $n=70$.
Thus, the GEK model is not feasible when the number of sampling points is high.

In the next section, we increase the number of sampling points on the same analytic function and compare the GE-KPLS$m$ models for $m=1,\dots,5$ to the kriging, KPLS and KPLSK models

\subsection{Numerical results for the analytical functions: second study}
\label{sec:study2}

\begin{figure}[H]
\centering
\includegraphics[width=\linewidth]{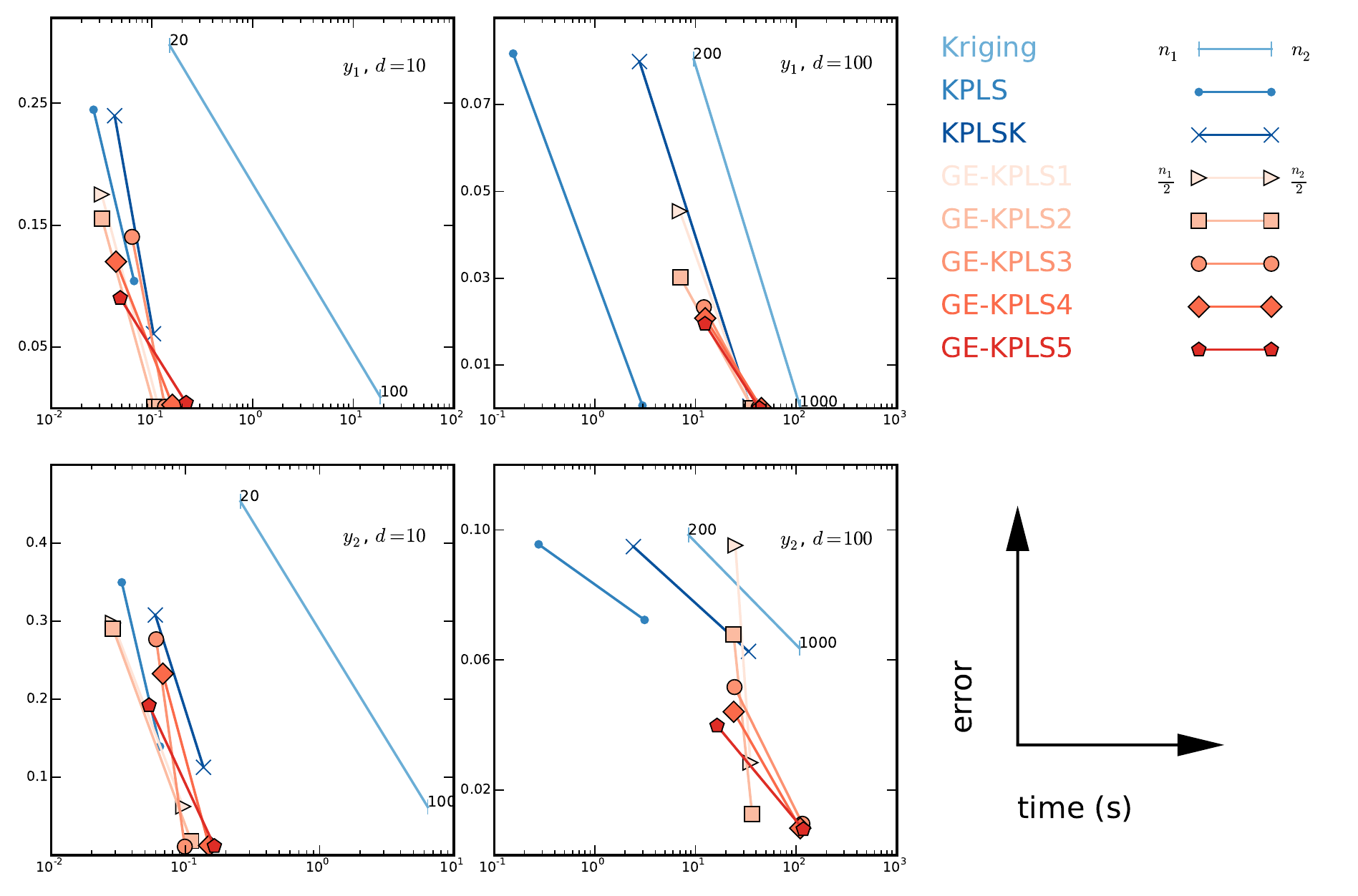}
\caption{\label{res:anal}Summary of results for all models and analytical problems, based on 10 trials for each case.
The models are color coded as shown legend (upper right).
The best trade-off between time and error is always obtained by a GE-KPLS model.}
\label{fig:analytic}
\end{figure}

For the second study, we use again the analytical functions $y_1$ and $y_2$ respectively given by Equations~\eqref{eq:y1} and~\eqref{eq:y1}.
For each case, the number of sampling points, $n$, used is equal to $kd$, with \mbox{$k=2,10$}, for the kriging, KPLS and KPLSK models, while $\frac{n}{2}$ sampling points are used for the GE-KPLS$m$ models.
To analyze the trade-off between the computational time and RE, we plot the computational time versus RE in Figure~\ref{res:anal}, where each line represents a given surrogate model using two different numbers of samples.
Each line has two points corresponding to $n_1 = 2d$ and $n_2=10d$ sampling points for the kriging, KPLS, and KPLSK models; and to $\frac{n_1}{2}$ and $\frac{n_2}{2}$ for the GE-KPLS$m$ models.
The models are color coded according to the legend in the upper right sequence of model names, starting with kriging through GE-KPLS5.
The rows in this grid show the results of different functions, starting with $y_1$ on the top, and ending with $y_2$ in the bottom.
The columns show the results of different dimensions, starting with $d=10$ on the left, and ending with $d=100$ dimensions on the right of the corresponding function.
More detailed numerical results for the mean of the RE and computational time are listed in Table~\ref{tab:reAnal} in Appendix~\ref{app:analCases}.

As we can see in Figure~\ref{res:anal}, adding $m$ extra points to the correlation matrix improves the accuracy of the results, and the best trade-off between time and error is always obtained given by a GE-KPLS model.
At the expense of a slight increase in computational time, increasing the number of extra points always yields to a lower error in almost all cases.
Indeed, the GE-KPLS5 yields to a lower error for all cases except for $y_1$ and $y_2$ with 10 dimensions and 50 sampling points, where the lowest error is obtained with the GE-KPLS2 and GE-KPLS3 models, respectively.
Thus, the number of extra points must be carefully selected.

We can evaluate the performance by either comparing the computational time required to achieve a certain level of accuracy, or by comparing the accuracy for a given computational time.
GE-KPLS1, for instance, provides an excellent compromise between error and computational time, as it is able to achieve a RE lower than 1\% under 0.1\,s for the $y_1$ function with 10 dimensions and 50 points.
Even better, the GE-KPLS5 yields a lower RE using 100 sampling points than the kriging model using 1000 sampling points (1.94\% vs. 2.96\%) for the $y_1$ function with $d=100$.
In this case, the computational time required to build a GE-KPLS3 is lower by a factor of 9 compared to the computational time needed by kriging (12.4\,s vs 109.6\,s).
In addition, the GE-KPLS method is able to avoid ill-conditioned correlation matrices by reducing the number of extra points through the PLS approach.

Thus, this second study confirms the efficiency of the GE-KPLS$m$ models and their ability to generate accurate models.

\subsection{Numerical results for the engineering functions: third study}
\label{sec:study3}

We now assess the performance of the GE-KPLS models on 8 engineering functions $\text{P}_1,\dotsc,\text{P}_8$ listed in Table~\ref{dimSampEng}.
The first three functions are the deflection, bending stress, and shear stress of welded beam problem~\citep{Deb1998}.
The fourth function considers the water flow rate through a borehole that is drilled from the ground surface through two aquifers~\citep{Morris1993}.
The Fifth function gives the position of a robot arm~\citep{An2001}.
The sixth function estimates the weight of a light aircraft wing~\citep{Forrester2008}.
P$_7$ and P$_8$ are, respectively, the weight and the lowest natural frequency of a torsion vibration problem~\citep{Wang2006}.
The number of dimensions for each of these problems varies from 2 to 15.
The detailed formulation for these problems is provided in Appendix~\ref{app:eng}.
In this study, we have intentionally chosen to cover a large engineering areas using a different number of dimensions and complexities, thus we can verify the generalization and the applicability of our approach.
To build the kriging, KPLS, and KPLSK models, we use two different number of sampling points, $n_1=2d$ and $n_2=10d$, for all problems except for $\text{P}_1$, $\text{P}_2$ and $\text{P}_3$ where $n_1=5d$ (see Table~\ref{dimSampEng} for more details).
Similarly, we use ${n_1}/{2}$ and ${n_2}/{2}$ sampling points for the GE-KPLS models.
We use GE-KPLS to construct surrogate models for these engineering functions, and compare our results to those obtained by kriging, KPLS, and KPLSK.
As in the analytical cases, we performed 10 trials for each case and used the same metrics of comparison: computational time and RE.
For our GE-KPLS surrogate model and as previously mentioned, we vary the number of extra points $m$ with $m=1,\dotsc,5$ and use one principal component for all problems, except for $\text{P}_1$, $\text{P}_2$ and $\text{P}_3$ where we use at most 2, 2 and 4 extra points, respectively.

\begin{figure}
\centering
\includegraphics[width=\linewidth]{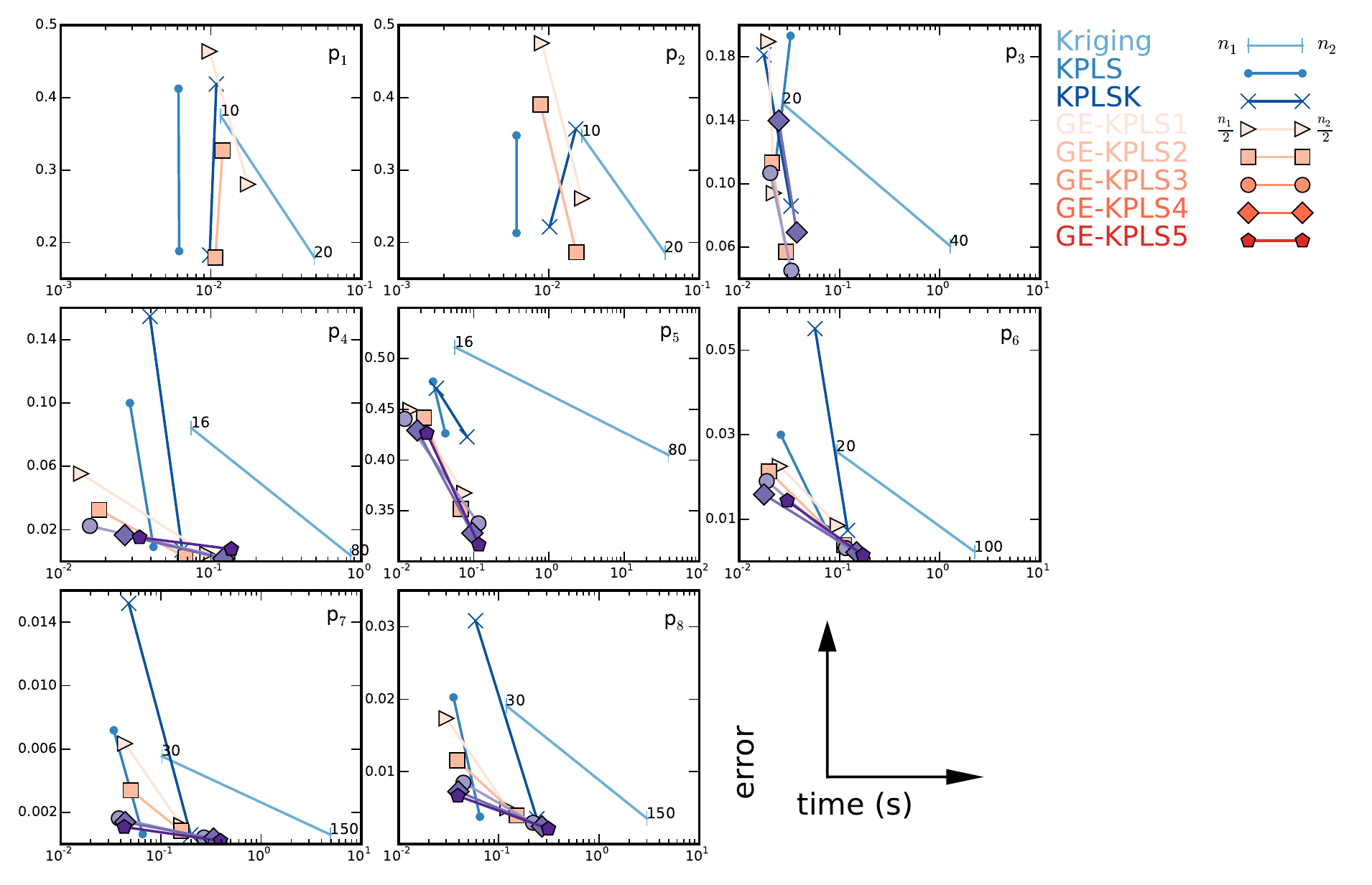}
\caption{\label{res:eng}Summary of results for all models and engineering problems, based on 10 trials for each case.
The models are color coded as shown in legend (upper right).
The best trade-off (time vs error) is always obtained by a GE-KPLS models.
}
\label{fig:eng}
\end{figure}

Figure~\ref{res:eng} shows the numerical results for the engineering functions.
As in the plots for the analytical cases, each line has two points corresponding to $n_1$ and $n_2$ sampling points for the kriging, KPLS, and KPLSK models; and to ${n_1}/{2}$ and ${n_2}/{2}$ for the GE-KPLS$m$ models.
The models are color coded according to the legend on the upper right.
This grid of plots shows the results of different problems, starting from P$_1$ on the top left and ending with P$_8$ on the bottom center.
The actual mean values of computational time and RE are given in the Table~\ref{tab:reEng} in Appendix~\ref{app:engCases}.

Overall, the GE-KPLS method yields a more accurate solution except for P$_2$.
For the P$_2$ function, GE-KPLS2 is almost as accurate than kriging (the model giving the best result) using 10 and 20 sampling points, respectively, with a relative error of 0.1866 for the former and 0.1859 for the latter.
All of the GE-KPLS results have either a lower computational time, lower error, or both, when compared to the kriging results for the same sampling cost.
This means that despite the augmented size of the GE-KPLS correlation matrices, the computational time required to build these models is lower than the kriging model.
The efficiency of GE-KPLS is due to the reduced number of hyperparameters that need to be estimated, which is 1 in our case, compared to $d$ hyperparameters for kriging.
In addition, our strategy of how using the PLS coefficients to rescale the correlation matrix results in better accuracy.
For example, we need only 0.12\,s for P$_5$ with 80 sampling points to build a GE-KPLS5 model with a relative error of 0.3165 compared to 38.9\,s for a kriging model (best error given the benchmark) with a relative error of 0.4050.

In Figure~\ref{res:eng}, we notice that the computational time required to train some models for P$_1$ and P$_2$ seems higher when decreasing the number of sampling points, like for example the KPLSK model for P$_2$.
This is due to the fast construction of the model for such problems with low dimensions, and the difference in computational time using the different number of sampling points is less than $10^{-3}$\,s.

This study shows that the GE-KPLS model is accurate and computationally efficient for different engineering fields.
In addition, a given user is able to choose the best compromise between computational time and error.
One way to do it is to start by the construction of a GE-KPLS1 model.
Then, the user fixes a reasonable trade-off between the error and the computational time (guided by the first results), and subsequently adds more approximating points to achieve such compromise.
Another way to select $m$ is to define a threshold and to keep approximating points with a higher PLS coefficients.
We also note that the selection of the number of additional points $m$ should be carefully done by the user with regards to his final goal.
For example, if the user uses a surrogate model within an iterative optimization design process, it is better to select a GE-KPLS model with a relatively low number of approximating points, since new many sampling points would be added close to each other in a small region that quickly deteriorate the condition number of the correlation matrix.
In the other side, if the final goal is to construct an accurate surrogate model over the design space, the number of approximating points $m$ could be relatively high.

\section{Limitation of the GE-KPLS method}
\label{sec:limitation}

Despite the numerous advantages of the proposed method relative to established models in the literature, there are still some issues that the use must be concerned with.
The major issue with GE-KPLS, which is a common issue with most methods in the literature, is what values to use in certain model parameters.
One of these parameters is the step size of FOTA, that was first optimized when GE-KPLS is used in Sections~\ref{sec:study1}, \ref{sec:study2} and \ref{sec:study3}, is an important parameter that can influence the final results, as this parameter is very sensitive to the type of problem and the sampling points.

In terms of implementation, the current toolbox version to build GE-KPLS cannot handle problems with a large number of both dimensions and sampling points.
This issue is mainly due to the memory required during the inversion of the correlation matrix.
An approximation of this memory limit is given by
\(n = 30550d^{-0.427}.\)
This estimation is fit by an approximation of the tendency between $n$ and $d$ through Microsoft Excel.

\section{Conclusions}
\label{sec:concl}

We developed a novel approach that uses gradient information at the sampling points that builds accurate kriging surrogate models for high-dimensional problems efficiently.
The proposed approach differs from classical strategies, such as the indirect and direct gradient enhanced kriging in that we exploit the gradient information without dramatically increasing the size of the correlation matrix, and we reduce the number of hyperparameters.
We applied the PLS method on each sampling point and selected the most relevant approximating points to include into the correlation matrix given by the PLS information.
Through some elementary operations on the kernels, we accelerated the construction of the model by using the average of all computed PLS information to reduce the number of hyperparameters.
Thus, our approach scales well the number of independent variables by reducing the number of hyperparameters, and the number of sampling points by selecting only relevant approximating points.

To demonstrate the computational efficiency and the accuracy of the proposed model, we presented a series of comparisons for both analytic functions and engineering problems with different number of both dimensions and sampling points.
Three comparison studies were performed.
We first compared our approach using $m$ extra points for $m=1,\dots,5$ and $m=d$ to the indirect gradient enhanced kriging and ordinary kriging models.
With $m=d$, which is the same number of extra points used for gradient enhanced kriging, we showed the usefulness of the PLS method in terms of computational time and error.
The results of this study also demonstrated the effectiveness and accuracy of the GE-KPLS$m$ models for $m=1,\dots,5$ when compared to other models.
In some cases, the GE-KPLS model is over 3 times more accurate and over 3200 times faster than the indirect GEK model.
In the second study, we increased the number of sampling points for the same analytic functions to compare the GE-KPLS$m$ for $m=1,\dots,5$ with the kriging, KPLS, and KPLSK models.
This study confirmed the results obtained by the first study, and GE-KPLS shows excellent performance both in terms of the computational time and the relative error.
For the first function and compared to both kriging and KPLS models, the accuracy is an order of magnitude smaller.
The third study focused on 8 engineering functions using the GE-KPLS$m$ for $m=1,\dots,5$ and the kriging and KPLS(K) models.
The GE-KPLS models yielded more accurate models for 7 problems irrespective of the number of both dimensions and sampling points.
The improvement in terms of relative error provided by the GE-KPLS model is up to 9\% in some cases.

GE-KPLS is able to freely manage the number of approximating points for avoiding ill-conditioned matrices with an important gain in terms of computational time and accuracy.
This kind of flexibility is very convenient in real applications especially when the surrogate model is used within an iterative sampling method; e.g. optimization design.
Indeed, the user can reduce progressively the number of approximating points after a certain number of iterations for minimizing the risk of ill-conditioned problems, that is not available, or needs more sophisticated techniques, with standard gradient-enhanced kriging.
For an effective use of our method, we recommend a prior analysis of the step parameter.
Unfortunately, this parameter is a problem dependent and is impossible to guess in advance.
Actually, this is a common difficulty for all gradient-based methods.
Finally, All test functions and models used in this paper are available on https://github.com/SMTorg/SMT, and could be reproduced.

\appendix

\section{Definition of the engineering cases}
\label{app:eng}

The analytical expressions of engineering cases are given by
\subsection{\texorpdfstring{$\text{P}_1,\text{ P}_2 \text{ and P}_3$}{}}

The three responses are the deflection $\delta$, bending stress $\sigma$, and shear stress $\tau$ of a welded beam problem~\citep{Deb1998}, respectively.
\begin{equation*}
\text{P}_1:\,\delta=\frac{2.1952}{t^3b},
\end{equation*}
\begin{equation*}
\text{P}_2:\,\sigma=\frac{504000}{t^2b},
\end{equation*}
\begin{equation*}
\text{P}_3:\,\tau=\sqrt{\frac{\tau'^2+\tau''^2+l\tau'\tau''}{\sqrt{0.25\left(l^2+(h+t)^2\right)}}},
\end{equation*}
where
$$\tau'=\frac{6000}{\sqrt{2}hl}, \quad\tau''=\frac{6000(14+0.5l)\sqrt{0.25\left(l^2+(h+t)^2\right)}}{2\left[0.707hl\left(\frac{l^2}{12}+0.25(h+t)^2\right)\right]}$$
and
\begin{table}[H]
\centering
\begin{tabular}{@{}cl@{}}
\toprule
Input variables&Range\\
\midrule
$h$&$[0.125,1]$\\
$b$&$[0.1,1]$\\
$l,t$&$[5,10]$\\
\bottomrule
\end{tabular}
\end{table}

\subsection{\texorpdfstring{$\text{P}_4$}{}}
This problem characterizes the flow of water through a borehole that is drilled from the ground surface through two aquifers~\citep{Morris1993}.
The water flow rate (m$^3$/yr) is given by
\begin{equation*}
\text{P}_4:\,y=\frac{2\pi T_u\left(H_u-H_l\right)}{\ln\left(\frac{r}{r_w}\right)\left[1+\frac{2LT_u}{\ln\left(\frac{r}{r_w}\right)r_w^2K_w}+\frac{T_u}{T_l}\right]},
\end{equation*}
where
\begin{table}[H]
\centering
\begin{tabular}{@{}clcl@{}}
\toprule
Input variables&Range&Input variables&Range\\
\midrule
$r_w$&$[0.05,0.15]$&$r$&$[100,50000]$\\
$T_u$&$[63070,115600]$&$H_u$&$[990,1110]$\\
$T_l$&$[63.1,116]$&$H_l$&$[700,820]$\\
$L$&$[1120,1680]$&$K_w$&$[9855,12045]$\\
\bottomrule
\end{tabular}
\end{table}

\subsection{\texorpdfstring{$\text{P}_5$}{}}
This function represents the position of a robot arm given by~\citep{An2001}
\begin{equation*}
\text{P}_5:\,y=\sqrt{\left(\sum\limits_{i=1}^4L_i\cos\left(\sum\limits_{j=1}^i\theta_j\right)\right)^2+\left(\sum\limits_{i=1}^4L_i\sin\left(\sum\limits_{j=1}^i\theta_j\right)\right)^2},
\end{equation*}
where
\begin{table}[H]
\centering
\begin{tabular}{@{}cl@{}}
\toprule
Input variables&Range\\
\midrule
$L_i$&$[0,1]$\\
$\theta_j$&$[0,2\pi]$\\
\bottomrule
\end{tabular}
\end{table}

\subsection{\texorpdfstring{$\text{P}_6$}{}}
This consists in an estimate of the weight of a light aircraft wing, given by~\citep{Forrester2008}
\begin{equation*}
\text{P}_6:\,y=0.036S_w^{0.758}W_{fw}^{0.0035}\left(\frac{A}{\cos^2\Lambda}\right)q^{0.006}\lambda^{0.04}\left(\frac{100tc}{\cos\Lambda}\right)^{-0.3}\left(N_zW_{dg}\right)^{0.49}+S_wW_p,
\end{equation*}
where
\begin{table}[H]
\centering
\begin{tabular}{@{}clcl@{}}
\toprule
Input variables&Range&Input variables&Range\\
\midrule
$S_w$&$[150,200]$&$W_{fw}$&$[220,300]$\\
$A$&$[6,10]$&$\Lambda$&$[-10,10]$\\
$q$&$[16,45]$&$\lambda$&$[0.5,1]$\\
$tc$&$[0.08,0.18]$&$N_z$&$[2.5,6]$\\
$W_{dg}$&$[1700,2500]$&$W_p$&$[0.025,0.08]$\\
\bottomrule
\end{tabular}
\end{table}

\subsection{\texorpdfstring{$\text{P}_7\text{ and P}_8$}{}}
The two quantities of interest in this problem are the weight and the lowest natural frequency of a torsion vibration problem, given by~\citep{Wang2006}
\begin{equation*}
\text{P}_7:\,y=\sum\limits_{i=1}^3\lambda_i\pi L_i \left(\frac{d_i}{2}\right)^2+\sum\limits_{j=1}^2\rho_j\pi T_j \left(\frac{D_j}{2}\right)^2,
\end{equation*}
\begin{equation*}
\text{P}_8:\,y=\frac{\sqrt{\frac{-b-\sqrt{b^2-4c}}{2}}}{2\pi},
\end{equation*}
where \\
$$K_i=\frac{\pi G_id_i}{32L_i},\quad M_j=\frac{\rho_j \pi t_jD_j}{4g},$$
$$J_j=0.5M_j\frac{D_j}{2},$$
$$b=-\left(\frac{K_1+K2}{J_1}+\frac{K_2+K3}{J_2}\right),$$
$$c=\frac{K_1K_2+K_2K_3+K_3K_1}{J_1J_2},$$
and
\begin{table}[H]
\centering
\begin{tabular}{@{}clcl@{}}
\toprule
Input variables&Range&Input variables&Range\\
\midrule
$d_1$&$[1.8,2.2]$&$L_1$&$[9,11]$\\
$G_1$&$[105300000,128700000]$&$\lambda_1$&$[0.252,0.308]$\\
$d_2$&$[1.638,2.002]$&$L_2$&$[10.8,13.2]$\\
$G_2$&$[5580000,6820000]$&$\lambda_2$&$[0.144,0.176]$\\
$d_3$&$[2.025,2.475]$&$L_3$&$[7.2,8.8]$\\
$G_3$&$[3510000,4290000]$&$\lambda_3$&$[0.09,0.11]$\\
$D_1$&$[10.8,13.2]$&$t_1$&$[2.7,3.3]$\\
$\rho_1$&$[0.252,0.308]$&$D_2$&$[12.6,15.4]$\\
$t_2$&$[3.6,4.4]$&$\rho_1$&$[0.09,0.11]$\\
\bottomrule
\end{tabular}
\end{table}

\section{Results of the analytical cases}
\label{app:analCases}

\begin{table}[H]
\caption{Mean of the error values (upper table) and computational times (lower table) for kriging, KPLS, KPLSK, and GE-KPLS$m$ for $m=1,\cdots,5$ based on 10 trials.
The best values are highlighted in bold blue type.}
\centering
{\scriptsize
\begin{tabular}{@{}cclcccccccc@{}}
\toprule
&$d$&$n$--$\frac{n}{2}$&kriging&KPLS&KPLSK&GE-KPLS1&GE-KPLS2&GE-KPLS3&GE-KPLS4&GE-KPLS5\\
\midrule
$y_1$&10&100--50&0.0092&0.1043&0.1134&0.0020&\color{blue}\bf 0.0011&0.0013&0.0029&0.0044\\
&&20--10&0.2976&0.2563&0.2555&0.1752&0.1556&0.1405&0.1201&\color{blue}\bf 0.0903\\
&100&1000--500&0.0296&0.0615&0.0562&0.0081&0.0051&0.0041&0.0032&\color{blue}\bf 0.0026\\
&&200--100&0.0805&0.0818&0.0817&0.0454&0.0302&0.0233&0.0207&\color{blue}\bf 0.0194\\
$y_2$&10&100--50&0.0618&0.1393&0.1475&0.0623&0.0182&\color{blue}\bf 0.0110&0.0123&0.0116\\
&&20--10&0.4532&0.3787&0.3297&0.2976&0.2903&0.2766&0.2325&\color{blue}\bf 0.1920\\
&100&1000--500&0.0637&0.0723&0.0695&0.0285&0.0126&0.0097&0.0083&\color{blue}\bf 0.0080\\
&&200--100&0.0984&0.0956&0.0950&0.0952&0.0678&0.0517&0.0441&\color{blue}\bf 0.0398\\
\midrule
\midrule
$y_1$&10&100--50&18.57&\color{blue}\bf 0.07&0.09&0.12&0.10&0.13&0.16&0.22\\
&&20--10&0.15&\color{blue} \bf 0.02&0.04&0.03&0.03&0.06&0.04&0.05\\
&100&1000--500&109.59&\color{blue}\bf 2.97&33.59&35.31&36.91&42.23&45.11&43.04\\
&&200--100&9.58&\color{blue}\bf 0.15&2.59&6.99&7.06&12.12&12.49&12.42\\
$y_2$&10&100--50&6.39&\color{blue}\bf 0.06&0.11&0.09&0.11&0.09&0.15&0.16\\
&&20--10&0.26&\color{blue}\bf 0.02&0.05&0.03&0.03&0.06&0.07&0.05\\
&100&1000--500&107.70&\color{blue}\bf 3.12&33.24&35.08&36.49&115.24&109.76&117.65\\
&&200--100&8.52&\color{blue}\bf 0.28&2.39&24.92&23.89&24.40&23.95&16.37\\
\bottomrule
\label{tab:reAnal}
\end{tabular}}
\end{table}

\section{Results of the engineering cases}
\label{app:engCases}
\begin{table}[H]
\caption{Mean of the error values (upper table) and computational times (lower table) for kriging, KPLS, KPLSK, and GE-KPLS$m$ for $m=1,\cdots,5$ based on 10 trials.
The best values are highlighted in bold blue type.}
\centering
{\scriptsize
\begin{tabular}{@{}cclccccccccc@{}}
\toprule
&$d$&$n$--$\frac{n}{2}$&kriging&KPLS&KPLSK&GE-KPLS1&GE-KPLS2&GE-KPLS3&GE-KPLS4&GE-KPLS5\\
\midrule
$\text{P}_1$&2&20--10&0.1796&0.1881&0.1827&0.2804&\color{blue}\bf 0.1791&--&--&--\\
&&10--5&0.3747&0.4124&0.4188&0.4639&\color{blue}\bf 0.3268&--&--&--\\
$\text{P}_2$&2&20--10&\color{blue}\bf 0.1859&0.2132&0.2216&0.2609&0.1866&--&--&--\\
&&10--5&\color{blue}\bf 0.3474&0.3478&0.3567&0.4751&0.3904&--&--&--\\
$\text{P}_3$&4&40--20&0.0607&0.0981&0.0906&0.0940&0.0571&\color{blue}\bf 0.0451&0.0692&--\\
&&20--10&0.1504&0.1933&0.1813&0.1897&0.1132&\color{blue}\bf 0.1067&0.1398&--\\
$\text{P}_4$&8&80--40&0.0037&0.0118&0.0091&0.0046&0.0026&0.0019&\color{blue}\bf 0.0017&0.0078\\
&&16--8&8.41&0.0999&0.1546&0.0554&0.0325&0.0224&0.0166&\color{blue}\bf 0.0151\\
$\text{P}_5$&8&80--40&0.4050&0.4347&0.4296&0.3674&0.3518&0.3376&0.3278&\color{blue}\bf 0.3165\\
&&16--8&0.5114&0.4773&0.4707&0.4493&0.4418&0.4405&0.4292&\color{blue}\bf 0.4264\\
$\text{P}_6$&10&100--50&0.0023&0.0101&0.0086&0.0085&0.0039&0.0031&0.0022&\color{blue}\bf 0.0015\\
&&20--10&0.0260&0.0300&0.0551&0.0225&0.0213&0.0190&0.0158&\color{blue}\bf 0.0144\\
$\text{P}_7$&15&150--75&0.0006&0.0008&0.0007&0.0012&0.0008&0.0004&0.0003&\color{blue}\bf 0.0002\\
&&30--15&0.0055&0.0072&0.0152&0.0063&0.0034&0.0016&0.0014&\color{blue}\bf 0.0011\\
$\text{P}_8$&15&150--75&0.0035&0.0041&0.0037&0.0050&0.0040&0.0029&0.0024&\color{blue}\bf 0.0021\\
&&30--15&0.0191&0.0202&0.0308&0.0173&0.0115&0.0085&0.0073&\color{blue}\bf 0.0067\\
\midrule
\midrule
$\text{P}_1$&2&20--10&0.05&\color{blue}\bf 0.006&0.01&0.02&0.01&--&--&--\\
&&10--5&0.01&\color{blue}\bf 0.006&0.01&0.01&0.01&--&--&--\\
$\text{P}_2$&2&20--10&0.06&\color{blue}\bf 0.006&0.01&0.02&0.01&--&--&--\\
&&10--5&0.02&\color{blue}\bf 0.006&0.01&0.01&0.01&--&--&--\\
$\text{P}_3$&4&40--20&1.27&\color{blue}\bf 0.01&0.02&0.02&0.03&0.03&0.04&--\\
&&20--10&0.03&\color{blue}\bf 0.01&0.02&0.02&0.02&0.02&0.02&--\\
$\text{P}_4$&8&80--40&0.85&\color{blue}\bf 0.03&0.07&0.09&0.07&0.13&0.12&0.14\\
&&16--8&0.07&0.03&0.04&\color{blue}\bf 0.01&0.02&0.02&0.03&0.03\\
$\text{P}_5$&8&80--40&38.90&\color{blue}\bf 0.03&0.084&0.07&0.07&0.12&0.10&0.12\\
&&16--8&0.06&0.03&0.03&\color{blue}\bf 0.01&0.02&\color{blue}\bf 0.01&0.02&0.02\\
$\text{P}_6$&10&100--50&2.23&\color{blue}\bf 0.04&0.10&0.10&0.11&0.11&0.15&0.17\\
&&20--10&0.09&0.03&0.06&0.02&0.02&0.02&\color{blue}\bf 0.018&0.03\\
$\text{P}_7$&15&150--75&4.92&\color{blue}\bf 0.05&0.19&0.16&0.16&0.27&0.34&0.39\\
&&30--15&0.10&\color{blue}\bf 0.03&0.05&0.04&0.05&0.27&0.34&0.39\\
$\text{P}_8$&15&150--75&3.01&\color{blue}\bf 0.05&0.18&0.12&0.15&0.22&0.27&0.31\\
&&30--15&0.12&\color{blue}\bf 0.03&0.06&\color{blue}\bf 0.03&0.04&0.04&0.04&0.04\\
\bottomrule
\label{tab:reEng}
\end{tabular}}
\end{table}

\bibliographystyle{abbrvnat}
\bibliography{biblio}
\end{document}